\pdfoutput=1

\documentclass[11pt]{article}
\usepackage[final]{EMNLP2023}

\usepackage{times}
\usepackage{latexsym}
\usepackage{graphicx}

\usepackage[T1]{fontenc}

\usepackage[utf8]{inputenc}

\usepackage{microtype}

\usepackage{inconsolata}
\usepackage{tabularx}
\usepackage{xcolor}
\usepackage{algorithmicx}
\usepackage{algorithm}
\usepackage[noend]{algpseudocode}
\usepackage{booktabs}
\usepackage{amsmath}
\newcommand{\todo}[1]{\textcolor{red}{\small [TODO]}}

\title{Towards Harmful Erotic Content Detection through Coreference-Driven Contextual Analysis \\
\vspace{10pt}
\small\textcolor{orange}{WARNING: This paper contains material of a sensitive nature.}}

\author{Inez Okulska \and Emilia Wiśnios \\
\texttt{\{inez.okulska,emilia.wisnios\}@nask.pl} \\
NASK National Research Institute}

\begin{document}
\maketitle
\begin{abstract}
Adult content detection still poses a great challenge for automation. Existing classifiers primarily focus on distinguishing between erotic and non-erotic texts. However, they often need more nuance in assessing the potential harm. Unfortunately, the content of this nature falls beyond the reach of generative models due to its potentially harmful nature. Ethical restrictions prohibit large language models (LLMs) from analyzing and classifying harmful erotics, let alone generating them to create synthetic datasets for other neural models. In such instances where data is scarce and challenging, a thorough analysis of the structure of such texts rather than a large model may offer a viable solution. Especially given that harmful erotic narratives, despite appearing similar to harmless ones, usually reveal their harmful nature first through contextual information hidden in the non-sexual parts of the narrative. 

This paper introduces a hybrid neural and rule-based context-aware system that leverages coreference resolution to identify harmful contextual cues in erotic content. Collaborating with professional moderators, we compiled a dataset and developed a classifier capable of distinguishing harmful from non-harmful erotic content. Our hybrid model, tested on Polish text, demonstrates a promising accuracy of 84\% and a recall of 80\%. Models based on RoBERTa and Longformer without explicit usage of coreference chains achieved significantly weaker results, underscoring the importance of coreference resolution in detecting such nuanced content as harmful erotics. This approach also offers the potential for enhanced visual explainability, supporting moderators in evaluating predictions and taking necessary actions to address harmful content. 
\end{abstract}

\section{Introduction}

The identification of harmful content represents a fundamental application of Natural Language Processing (NLP) methods on the internet. Such harmful content encompasses various forms, including hate speech, offensive material, misinformation, and graphic content. Among these, harmful erotic narratives present a particularly sensitive challenge. However, \textbf{general adult content detection models primarily focus on distinguishing between non-erotic and erotic texts without nuances in terms of their potential harm}. This is particularly challenging given that the parts describing sexual encounters often appear quite similar in most narratives. \textbf{It is the contextual information and additional details describing the individuals involved in or subjected to these sexual actions that ultimately reveal their harmful nature}.

The sentence \textit{"He made love with her"} is a common example of an sexual-related sentence that falls under adult content classification but is generally harmless. However, if the model can detect that elsewhere in the text, at some distance, there is a hint that the term \textit{'her'} refers to a minor or a sibling, it can then raise awareness among readers or, even better, alert moderators to the potentially harmful nature of the text. \textbf{And that distant reference, that subtle hint, is precisely what coreference resolution is designed for – to comprehend the semantic chains throughout the narrative}.

In collaboration with moderators from an institution serving as part of a national Incidence Response Team, our research identified a gray area of harmful erotic content that could gain from a coreference-driven contextual analysis. While not yet illegal in certain jurisdictions, this content has the potential to inflict significant harm, particularly on younger or more vulnerable readers. 

In this paper, we propose a hybrid context-aware system, using neural and rule-based components, for harmful content detection. It utilizes the coreference module for Polish spaCy model \cite{tuora2019integrating} to find interrelations between potentially hamrful contextual cues and sex-related parts. While several erotic content classifiers already exist, there is a conspicuous absence of harmful erotic content classifiers, particularly in languages like Polish. Although our proposed approach has been tested on Polish text, it can readily adapt to other languages, given the availability of BERT-based models and coreference resolution tools for those languages (which is the case for example for the English language).

The primary challenge in developing such a classifier is the scarcity of data. As this form of narrative content straddles the line between illegality and deviance, sourcing and collecting suitable training data poses a formidable obstacle. Unfortunately, the content of this nature falls beyond the reach of generative models due to its potentially harmful nature. \textbf{Ethical restrictions prohibit large language models (LLMs) from analyzing and classifying harmful erotics, let alone generating them to create synthetic datasets for other neural models}.

In our research, we assembled, together with the professional moderators, a modest dataset of 164 text samples, meticulously curated and flagged by them. While this dataset remains insufficient for training a robust classifier, it has proven adequate for extracting the harmful contextual features for the hybrid content classifier. 

As a result, \textbf{we introduce a hybrid model capable of distinguishing between non-harmful and harmful erotic content by leveraging both sexual content predictions and contextual cues}. Our experiments, using real-life examples assessed by professional moderators, demonstrate the promise of this approach, achieving an accuracy of 84\% and a recall of 80\%. Furthermore, the model's high potential for explainability, thanks to its hybrid coreference-driven architecture, holds great significance for human moderators. They require this level of understanding to evaluate each prediction, make informed decisions, and ultimately classify the text, potentially taking actions such as removing the content from the web or contacting the authorities.


\section{Related Work}

The task of coreference resolution for the Polish language has been gaining attention for many years \cite{ogrodniczuk2014coreference, niton2018deep}. In 2022, it was also one the subjects of the Shared Task at CRAC 2022 \cite{saputa2022coreference}. There are free coreference tools available -- a rule-based and a statistical resolution tool\footnote{\url{http://zil.ipipan.waw.pl/PolishCoreferenceTools}}. Both utilizing the Polish Coreference Corpus \cite{ogrodniczuk2016polish}. Applications to which coreference resolution in Polish has been applied include document summarization \cite{kopec2019three} and information extraction \cite{kaczmarek2015evaluation}. In the German language, it has been used for drama analysis \cite{pagel2020gerdracor}, and in English, for the analysis of medical interviews \cite{uzuner2012evaluating}.

The issue of gender bias in the Polish language concerning the detection of coreference chains was also studied \cite{zhu2016kobani, zhao2018gender, kocmi2020gender}).

As for the detection of harmful content, the harmfulness of which becomes evident only in the context of the interrelationships between mentions, has not been the subject of research so far. This applies primarily to the Polish language, but also, to the best of our knowledge, to other languages.

Efforts in the field of automated Child Sexual Abuse Material (CSAM) detection have predominantly focused on identifying harmful images and videos \cite{LEE2020301022}. The use of actual CSAM material for model training is constrained by significant legal and ethical complexities. Consequently, researchers have explored alternative approaches, with an emphasis on metadata and filename detection \cite{pereira2021metadatabased}.

While textual CSAM content has garnered relatively less attention, Natural Language Processing (NLP) methods and stylometric techniques, such as author profiling, have been adapted for online child grooming detection \cite{BORJ2023110039}. Emil Fleron, utilizing a dataset of abuse forum connections from the 2017 Freedom Hosting 2 dark net leak, investigated how supervised machine learning, relying solely on text data, can identify posts linked to CSAM distribution \cite{Fleron}. Text mining techniques have also been applied to the examination of medical documentation related to child abuse in the Netherlands \cite{Amrit}. In a different context, NLP-based methods have been employed to detect sexual/erotic content in user-generated online texts, aimed at filtering out content inappropriate for minors \cite{barrientos_2020}.

Recent studies have ventured into sentence-level pornographic content detection in Chinese and English datasets comprising novels and stories \cite{song2021}. However, these approaches are primarily designed to identify adult content, often neglecting the consideration of its harmful or non-harmful nature, and notably, none of them incorporate methods using coreference resolution.

In conclusion, the automated detection of harmful erotic narratives requires further development and investigation, and the incorporation of a coreference-based method represents a novel contribution to this field.

\section{Data Collection and Preprocessing}

Our hybrid coreference-driven model for harmful erotic content detection relies on two distinct datasets to facilitate and evaluate its functioning in Polish language: a set of sentences describing sexual encounters, called Sexual Sentences Dataset, and the collection of actual harmful erotic narratives, called Harmful Erotic Full-Text Dataset. 

\subsection{Sexual Sentences Dataset}\label{subsec:sentences_dataset}
The dataset comprises approximately 28000 sentences tokenized with NLTK library and selected for binary classification of sexual content with 5865 for class neutral and 22135 for sexual. We intentionally sampled and shuffled these sentences to obscure any contextual cues between adjacent sentences. This deliberate choice prevents the leakage of contextual information from one sentence to the next. For instance, consider the pair of sentences: "He touched her naked skin in a very intimate way. He could see that she loved it." When presented in an unshuffled narrative, the second sentence often led to false positive label 'sexual' due to prior knowledge, although it does not indicate any sexual activity itself. However, by annotating each sentence individually, such ambiguities were minimized.

Each sentence underwent manual labeling by three human annotators, final label was assigned as a result of majority voting. The percentage agreement was 87\%. More details regarding annotation process is presented in the Appendix \ref{sec:annotation}. 

These sentences were sourced from both non-professional and professional narratives gathered from a diverse array of online sources. These sources included web services specializing in short stories contributed by various authors, spanning categories such as "love," "life," "friendship," and "erotic." 
We did not perform any additional text preprocessing. 

\subsection{Harmful Erotic Full-Text Dataset}

The second dataset served for the main task of the general detection of harmful erotic narratives, and has been split into training, test and validation set. The first one -- encompassing 308 samples, has been used for analysis of coreference structures emerging in this type of narrative texts. The same dataset was used for training baseline models (see Section \ref{sec:experiments}). The test set made of 78 samples was used for evaluating the performance of baseline models (Appendix \ref{app:parameters}). The experiments proving the performance of the presented method in numbers, have been run on 164 previously unseen samples. 

The harmful class within this dataset comprises text content collected by automated scrapers commissioned by a legal institution tasked with addressing cyber incidents of this nature. Manual classification of these texts was carried out by professionally trained moderators as part of their daily responsibilities.

In the beginning, the full-text corpus exclusively comprised non-professional narratives categorized by moderators under the CSAM (Child Sexual Abuse Material) classification. However, as we collaborated on developing automated classification algorithms, we identified a 'gray zone' of texts. These texts initially fell outside the strict confines of the CSAM definition but were nonetheless deemed disturbing and deviant by both professional and non-professional annotators, including members of the machine learning team.

As a result, we made the decision to expand the category beyond CSAM to encompass harmful-erotic content. This broader category includes all samples describing sexual relations involving young individuals marked by significant age and authority differentials, such as teacher-student dynamics, as well as various forms of incestuous narratives. This expansion specifically encompasses narratives where the focus lies on the sexual excitement induced by family relationships and/or the innocence of a young person, even extending to scenarios involving cousins.

The gathered text samples were tokenized using the same SpaCy model that was employed during the training of the binary sexual sentence classifier. No additional preprocessing steps were applied either before or after tokenization, ensuring the classifier's robustness in handling the natural online presence of such content.

\section{RoBERTa-Based Sexual Sentence Classifier}

The neural part of our hybrid approach relies on the sentence-level sexual sentence transformer classifier. It consists of RoBERTa base\footnote{\url{https://huggingface.co/sdadas/polish-roberta-base-v2}} and additional linear layers with dropout and ReLU activation build on top of the RoBERTa hidden state. The final version utilizes only the base version of the model, as our initial experiments have shown that bigger model does not improve the classification results significantly. 

The Sexual Sentence Dataset (as described in Section \ref{subsec:sentences_dataset}) was divided into train (22400) and test (5600) with similar distributions of the classes in both datasets, being approximately 3:1 (non-sexual:sexual).

In its final architecture the model utilizes three linear layers were (with sizes: 768, 512, 256, 1). Dropout probability was equal 0.2. The model was trained on 5 epochs (effect of early stopping based on the validation loss), using Adam as the optimizer with the learning rate of 1e-5. 
Table \ref{tab:classification_metrics} shows the results.

\begin{table}[ht]
  \caption{Classification Report for validation set. "Prec." = precision, "Rec." = recall, "Sup" = support.}
  \label{tab:classification_metrics}
  \begin{tabularx}{\columnwidth}{Xcccc}
    \toprule
     & \textbf{Prec.} & \textbf{Rec.} & \textbf{F1} & \textbf{Sup.} \\
    \textbf{Non-Sexual} & 0.96 & 0.96 & 0.96 & 4427 \\
    \textbf{Sexual} & 0.84 & 0.83 & 0.84 & 1173 \\
    \text{Accuracy} &  &  & 0.93 & 5600 \\
    \text{Macro avg} & 0.90 & 0.89 & 0.90 & 5600 \\
    \text{Weighted avg} & 0.93 & 0.93 & 0.93 & 5600 \\
    \bottomrule
  \end{tabularx}
\end{table}

\section{Coreference Resolution for Contextual Analysis}

Our choice of the coreference resolution method is driven by the recognition that it is the context or scene within these narratives that typically distinguishes harmless erotica from potentially harmful variants. Detailed descriptions of the actors or objects involved in the sexual actions often reside in separate sentences from those describing the actions themselves. The presented method combines information from sentences classified as sexual, i.e., those describing sexual activities, with contextual information linked to individuals mentioned in these sentences through a coreference chain as shown in the Algorithm 1. We have decided to use a coreference model for the Polish language proposed by \citet{saputa-2022-coreference}, based on the HerBERT model \cite{mroczkowski-etal-2021-herbert}. This is an end-to-end model, conveniently integrated into the Polish spaCy model, that achieved an F1 score of 76.67 in the CRAC Shared Task 2022 \cite{crac-2022-crac} on the Polish test dataset.

First, each sentence receives prediction regarding whether they contain sexual content or not. For the list of all sexual sentences, position indices in the document are determined, ranging from the first to the last token for each sexual sentence. Subsequently, all detected coreference chains in the text are examined to determine if they contain significant contextual information based on the contextual features described in Section 6. If such elements are identified, mention positions are established and compared with the index ranges for the sexual sentences. This process checks whether a given potentially harmful cue refers to another word that is part of a sentence describing sexual activity. Grammatical dependencies are also examined concerning the verbal phrases in the sexual sentence.

Contextual elements in such chains can contain information directly — the same or synonymous noun forms, personal pronouns, or possessives. A crucial aspect of this analysis is the syntactic relationship, which allows us to determine whether, in the case of multiple references to individuals, they are indeed participating in the activities described in the sexual sentence.

\begin{figure}[ht]
\noindent
\begin{minipage}[t]{\columnwidth}
\begin{algorithm}[H]
\footnotesize
    \caption{Coreference resolution between harmful contextual features and sex-related sentences}
    \label{alg:winky}
    
\begin{tabular}{ c l }
    \textbf{Requires: } \\
    $\mathtt{doc}$ & The SpaCy Doc type \\
    $\mathtt{sentences}$ & List of input sentences \\
    $\mathtt{coref\_chains}$ & List of doc coreference chains \\
    $\mathtt{harm\_context\_feats}$ & List of contextual features \\ & (creating harmful context) \\
    $\mathtt{seuxal\_model}$ & RoBERTa-based sexual sentence \\ &  classifier
\\
\end{tabular}
\begin{algorithmic}[1]
\State $\mathtt{sexual\_ids} \gets [\:]$
\For{$\mathtt{sentence} \in \mathtt{sentences}$}
    \State $\mathtt{sexual\_content} \gets \Call{sexual\_model}{\mathtt{sentence}}$
    \State $\mathtt{sexual\_sent\_id} \gets \Call{sentence position in doc}{}$
    \If{$\mathtt{sexual\_content} >= 0.5$}
        \State $\mathtt{sexual\_ids}.\Call{append}{\mathtt{sexual\_sent\_id}}$
    \EndIf
\EndFor
\State

\State $\mathtt{harm\_context\_clusters} \gets [\:]$
\For{$\mathtt{chain} \in \mathtt{coref\_chains}$}
    \State $\mathtt{chain\_ids} \gets \Call{all doc positions in chain}{}$
    \For{$\mathtt{mention} \in \mathtt{chain}$}
    \If{$\mathtt{mention} \in \mathtt{harm\_context\_feats}$}
        \State
        $\mathtt{harm\_context\_ids}.\Call{append}{\mathtt{chain\_ids}}$
    \EndIf
    \If{$\mathtt{harm\_context\_ids} \in \mathtt{sexual\_ids}$}
        \State $\mathtt{harm\_context\_clusters}.\Call{append}{\mathtt{chain}}$
    \EndIf
    \EndFor
    \If{$\mathtt{harm\_context\_clusters}$}
    \State 
    $\mathtt{label}$ = harmful
\Else{}
    \State 
    $\mathtt{label}$ = non-harmful
\EndIf
\EndFor
\State
\Return{$\mathtt{label}$}
\end{algorithmic}
\end{algorithm}
\end{minipage}
\end{figure}

\begin{figure*}
    \centering
    \includegraphics[width=\textwidth]{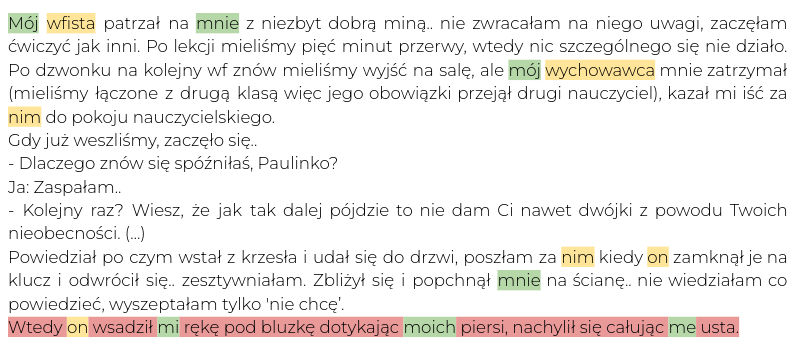}
    \caption{Example of visualization the contextual clues through coreference resolution. Marker colors: 1) red - sentence classified as sexual, 2) yellow - one coreference chain referring to the male person involved in the sexual action, 3) green - coreference chain referring to the subject of the sexual action.}
    \label{fig:enter-label}
\end{figure*}

Figure \ref{fig:enter-label} illustrates how contextual cues are distributed, and how the coreference mechanism allows for their identification, thereby distinguishing harmful from non-harmful erotic content, as well as shows the visualization potential of this approach. The provided example describes an erotic situation. The text highlighted in red has received a "sexual" label from the binary neural model because it describes sexual activities. However, these words are inherently neutral:  

\textit{'Wtedy on wsadził mi rękę pod bluzkę dotykając moich piersi, nachylił się całując me usta'} ('Then he slid his hand under my shirt, touching my breasts, leaned in to kiss my lips'). 

However, thanks to coreference resolution, it becomes evident that the \textit{'he'} in the sexual sentence is part of a longer chain, which allows us to decipher that it refers to a "guardian" and simultaneously a "P.E. teacher," indicating a physical education teacher.

On its own, the word \textit{'nauczyciel'} ('teacher') can merely serve as a hint about the profession, which is still insufficient to unequivocally classify the content as harmful. However, the second chain leaves no doubt. The analysis of the syntactic dependency in the highlighted sexual sentence demonstrates that the person who is subjected to the sexual activity 'Wtedy on wsadził \textit{mi} rękę pod bluzkę' ('he slipped his hand under \textit{my} shirt') is linked in a single chain with the descriptions '\textit{my} guardian' and '\textit{my} P.E. teacher,' unequivocally suggesting that this person is a student and we are dealing with a harmful sexual relationship between a school teacher and his female student.

Additionally, context elements related to the actors participating in sexual sentences but not directly describing the actors themselves, such as elements of their clothing or body parts, were examined. As described in Section \ref{sec:contextual_features}, the analysis of the training set revealed the unique presence of certain terms. Therefore, we decided to include them in the set of potentially harmful contextual clues that are detected in the coreference chain encompassing a sexual sentence.

 This also applies to various variations of age representation, where the presence of age-related terms was, as in other cases, linked through syntactic dependencies. If a sexual sentence contains subject or object descriptors of sexual activities that, through coreference chains, connect with another sentence containing age-related terms referring to the same person it is a clear indicator for the text label to be 'harmful'. For example, in the sexual sentence \textit{'He touched me'} and the \textit{'me'} connects with \textit{'I'} in another sentence, which is the subject of \textit{'I was 15 years old'}.  
    
\section{Contextual Features for Harmful Erotic Content Detection} \label{sec:contextual_features}

The coreference features have been automatically extracted and subjected to domain expert analysis. These features form the basis for the rule-based component of the hybrid model and allow for identifying specific elements of context relevant to the analysis of coreference chains related to the sentences describing sexual activities.

From the training set, all text samples involving sexual actions were selected, leaving two classes -- of harmful and non-harmful erotic narratives. The TF-IDF analysis was conducted on both of them separately, and based on it, the most frequently occurring words in their base forms (lemmas) were selected for each class. Only those belonging to the category 'noun' (part of speech) were filtered out from them. Then, among them, only those belonging to the 'person' semantic category were chosen, as the most crucial distinguishing element between non-harmful and harmful erotica is the set of actors participating in or being subject to sexual activities. This list ultimately includes 103 nouns. 

One of the most striking examples of differences is the word \textit{'mama'} ('mother'), which appeared 303 times in harmful erotica and only four times in the non-harmful class, and \textit{'syn'} ('son'), which appeared 151 times in harmful erotica and 0 times in the non-harmful class. This set of features mainly includes family members, teachers, and terms for children and young people in official, colloquial, and containing typical spelling errors.

Additionally, analyses of both lexicon distributions showed significant differences in the occurrence of additional elements describing scenery -- parts of clothing and body parts. In the case of the CSAM class, there was an overrepresentation of female clothing elements in the diminutive form (\textit{'spódniczka'}, 'little skirt', \textit{'staniczek', 'camisole')}, also in a characteristic form of non-professional, affective writing, which is not subject to editing, with spelling errors. Importantly, misspelled words do not undergo automatic lemmatization, so it was essential to include them in their original form, e.g.,: \textit{'soudniczke', 'sludniczke', 'spudnice', 'spudnicze', 'spudniczkach', 'spudniczki'} (there are all misspelled version of the word 'little skirt')

Words describing genitalia, characteristic of the harmful class (and not present in the non-harmful class), are vulgar forms that have undergone morphological diminution. On one hand, there is an "adult" term for intimate body parts suitable for sexual actions. On the other hand, there is some adjustment to underage participants (not suitable for sexual actions!) by using diminutive forms. Such a form (e.g., \textit{'kutasik'}, 'tiny little cock') is not encountered either in reference to adults (in that case, the diminutive would imply at least a derogatory, mocking attitude toward the described body part) or in neutral anatomical descriptions of children, where vulgar synonyms for official terms are not used, at most endearing ones (\textit{'siusiak'},'wee-wee').

As a result of the analysis of extracted nouns, it also turned out to be worthwhile to introduce a negative rule weakening the probability of prediction for the harmful erotic class based on words characteristic only for non-harmful erotica. In the harmful class, the words \textit{'mąż'} ('husband') and \textit{'żona'} ('wife') did not appear once (only one occurrence of \textit{'żoneczka'}, 'wifey'). In contrast, in the non-harmful erotic class, \textit{'żona'} and \textit{'mąż'} appear pretty often, 113 and 71, respectively.

Additional features detected in the text relate to age and include numerical, verbal, and verbal-numerical representations, taking into account typical forms of misspelling in terms of punctuation and spelling. The upper age limit, which is flagged, is 17 years. 

\section{Experiments and Results}\label{sec:experiments}

\begin{table*}[ht]
    \caption{Results from evaluation of all models on the test dataset made from 164 stories unseed in the training and validation phase. Details regarding parameters used for RoBERTa model and Longformer model are available in Appendix \ref{app:parameters}.}
    \label{tab:my_label}
    \begin{tabularx}{\textwidth}{Xcccc}
    \toprule
         \textbf{Model} &  \textbf{Recall} &  \textbf{Precision} &  \textbf{F1} & \textbf{Accuracy}\\ 
         RoBERTa base fine-tuned for 10 epoch&  91\%&  30\%&  63\%& 45\%\\ 
         RoBERTa base fine-tuned for 20 epoch&  70,5\%&  65\%&  68\%& 88\%\\ 
         Longformer&  82\%&  49\%&  61\%& 81\%\\ 
         \textbf{Coreference-Driven Hybrid Classifier}&  \textbf{80\%}&  \textbf{70,5\%}&  \textbf{75\%}& \textbf{84\%}\\ 
         Baseline Classifier (without coreference resolution)&  14\%&  100\%&  24\%& 77,5\%\\
         \bottomrule
    \end{tabularx}
\end{table*}

Detailed description of used RoBERTa model for sentence classification (first step of our classifier to identify sexual vs non-sexual content) is presented in Section \ref{tab:sentence-classification}. As for the full classification, the results were tested on independent real-life data, which consisted of 34 harmful stories and 130 non-harmful stories. 

For the full-text classification of harmful erotic content, we compared our proposed coreference-driven approach with RoBERTa base model\footnote{\url{https://huggingface.co/sdadas/polish-roberta-base-v2}} trained for 10 and for 20 epochs, the Longformer base\footnote{\url{https://huggingface.co/sdadas/polish-longformer-base-4096}}, and a baseline model. Similar to our proposed model, this baseline relies first on identifying sexual sentences with fine-tuned RoBERTa and then is looking for phrases suggesting harmful context (both the baseline and coreference-driven model utilize the same dictionaries and semantic rules as described in Section \ref{sec:contextual_features}). However, the main difference is that the baseline searches for these phrases exclusively in those sentences that have been identified (predicted) as sexual. In contrast, the coreference-driven model seeks contexts in sentences that are not necessary sexual per se (predicted in the first step as "neutral") but connected to sexual ones through the coreference chains. The results of both models demonstrate that the difference in searching for cues in a direct (without utilizing the coreference chains) and a broader context (with coreference chains) is crucial for capturing harmful content. 

As shown in Table 2, the RoBERTa base trained for ten epochs seems to be the best in the hunt for the harmful erotica with its 91\% recall. However, its focus on the harmful class let the precision drop to a disturbing rate of 30\%, meaning that 70\% of all non-harmful erotic stories would have been accused of containing some sort of deviation. The analogous issue can be observed in the case of the Langformer model, which offers high recall (82\%) but still a very low precision (49\%)). 

Thus, for the task of detecting harmful erotica the combination of recall, accuracy, and F1-Score is crucial for evaluating such a model instead of solely focusing on the highest recall. Detecting harmful content is a delicate matter since classifying a text with the "harmful" label may even cause legal actions. Therefore, leveraging high recall with high precision is significant in this case. Longer training of RoBERTa improved the precision significantly, but the recall fell by over 20 percentage points, which shows that this architecture cannot find the right balance for this case.

As already mentioned, one of the main challenges for detecting harmful erotic narratives is the collection of the training dataset. Most probably, the Longformer or the RoBERTa model could have shown more potential when presented with more training data. However, in this real-world case, it was necessary to find a working solution to overcome the problem of such a scarce dataset that was only possible to gather. Also given the fact that this kind of data cannot be effectively generated to enhance the dataset synthetically. 

The coreference-driven model achieves a satisfying recall (of 80\%) and accuracy (84\%), together with good precision and the best F1-Score (75\%). As the results show, the coreference-driven rules enhancing the neural sexual sentence classifier offer very promising alternative to the end-to-end models when the training data is lacking.

The results of full-text classification for the coreference-driven classifier are clearly dependent on the performance of the sexual sentence classifier. \textbf{Manual analysis of both false negatives and false positives of the coreference-driven hybrid model reveals that the majority of errors stem from an excessive or inaccurate classification of sentences as sexual.} Only in four cases of false negative, despite the accurate classification of sexual sentences, the syntactic relationships in the text proved to be too complex to unequivocally trace the connections between harmful contextual elements and the sexual actions and correctly label the text as harmful. This led us to believe that the proposed approach is perfectly worthy of consideration and further development, mainly focused on improvements of the RoBERTa-based sexual sentence classifier. 

\section{Discussion}

The main challenge in detecting harmful erotica lies in the fact that \textbf{merely identifying sexual sentences is insufficient. What makes a text genuinely harmful often requires a comprehensive reading to extract the information from a complex context, including entirely non-sexual sentences}. This is why the baseline model that searched for the contextual clues only in the direct sexual sentences failed significantly (with the recall of 14\%) in the overall detection of harmful content: the clues are usually located outside the sexual sentences. However, previous experiments have also shown that a simple full-text search for the clues (the keyphrases related to age, occupation, or family relationships) is also not enough. That is because these terms can appear in the text, but in a completely neutral context of world-building, unrelated to the sexual activity itself, an innocent part of the depicted world. \textbf{Hence, it is only through analysis using coreference resolution that we can search for and determine the true nature of the contextual clues and their relationships to the sexual content}.

\section{Limitations and Future Work}

\paragraph{Insufficient Harmful Data Availability.}
A notable limitation of this study lies in the limited availability of harmful data for thorough analysis. The acquisition and accessibility of datasets containing instances of harmful activities have posed significant challenges. To address this limitation, we intend to expand our data collection efforts in future research endeavors. Increasing the volume of available data is vital for enhancing the comprehensiveness and robustness of our analysis and findings.
\paragraph{The model for sexual sentence classification is far from perfect yet.} An essential component of the presented algorithm, despite the critical role of coreference, is the detection of sexual sentences within the text. If the model's prediction is incorrect, it can have a negative impact on the overall assessment of the text as harmful because each coreference chain is invariably linked to sentences classified as sexual. In the subsequent stages of the project, we plan to gather and annotate a more diverse set of data and then improve the quality of this classifier.

 \paragraph{Adoption of a Learning-Based Approach Over Rule-Based.}
The current contextual features were manually selected based on tf-idf analysis. With a larger corpus of available texts, it would be feasible to train a dedicated model capable of automatically detecting the presence of these features within the text. This shift toward a learning-based approach would enhance the system's adaptability and performance, as it could better capture intricate patterns and nuances within the data.

\section{Conclusion}

Addressing the demanding yet significant application of coreference resolution to harmful erotic content detection, we offer the following contributions:
\begin{enumerate}
    \item A first neural model fine-tuned solely for classifying sexual sentences in the Polish language, based on the RoBERTa model and trained on 28000 manually annotated sentences.
    \item A hybrid neural and rule-based model for detecting harmful erotic content, which leverages coreference resolution to extract necessary contextual clues. This way, it is capable of effectively distinguishing non-harmful from harmful erotic narratives.
    \item A visual explanation method for the model potentially highly beneficial for professional moderators involved in the detection of such texts in their work.
    \item Preliminary analysis of the issue of harmful erotica in the Polish language.
\end{enumerate}

\section*{Acknowledgments}
The results presented in this paper are a culmination of the research conducted as part of the project titled "APAKT -- A system responding to child safety threats in cyberspace with special emphasis on child pornography", generously funded by the National Center for Research and Development (NCBR) within the initiative CYBERSECIDENT/455132/III/NCBR/2020. 

We extend our heartfelt thanks to the team of moderators from 'Dyżurnet' who provided invaluable assistance throughout the research process.

\bibliography{custom}
\bibliographystyle{acl_natbib}

\appendix

\section{Annotation Details}
\label{sec:annotation}
In this section, we provide a comprehensive account of the annotation process, including the guidelines used, for classifying sentences as either sexual (1) or non-sexual (0) within the scope of our study. To minimize the potential influence of context, the sentences for annotation were intentionally shuffled. The process involved the participation of three annotators for each sentence, with the final label determined through a majority voting mechanism.

\subsection{Annotation Process}
\paragraph{Annotators.} Three human annotators were engaged for the task of sentence classification, each chosen for their proficiency in the target language and their prior experience with similar annotation tasks. These annotators were selected based on their ability to adhere to the annotation guidelines and their capacity to independently assess the sentences.

\paragraph{Majority Voting.} To maintain the robustness and objectivity of the annotation process, we employed a majority voting system. For each shuffled sentence, the three annotators independently assigned a label (1 for sexual or 0 for non-sexual). The final label for each sentence was determined through a majority vote. In instances of a tie, a consensus was reached through discussion among the annotators.

\subsection{Annotation Guidelines.}
Comprehensive and well-defined annotation guidelines were crucial to achieving consistency and accuracy in the annotation process. The following summarizes the key aspects of the annotation guidelines:
\paragraph{Sexual Sentence Definition.} A sexual sentence, as stipulated for this annotation task, is one that includes explicit content related to sexual activities or themes. Sentences depicting ordinary acts of affection such as kissing, holding hands, or hugging should not be classified as sexual. Annotators were instructed to focus on the presence of explicit or graphic language, descriptions of sexual acts, or content intended to discuss or explore sexual arousal as indicative of a sexual sentence.

\paragraph{Ambiguity and Context Independence.} Given that sentences were presented without context, annotators were instructed to assess each sentence independently. Ambiguity in the sexual nature of a sentence should be resolved based on the sentence's content alone. Annotators should not make assumptions or rely on contextual information that is not explicitly provided.

\paragraph{Consistency and Objectivity.} Annotators were encouraged to maintain a consistent approach throughout the annotation process and to avoid the introduction of personal biases. Classification should be solely based on the content of the sentence and its alignment with the provided definition of a sexual sentence.

\paragraph{Annotator Discussions.} In cases of uncertainty or disagreement among annotators, open discussions were encouraged to facilitate consensus and ensure the accuracy of the final label. Annotators were allowed to consult relevant reference materials or seek clarification from the research team to address any doubts.

\subsection{Inter-Annotator Agreement}

To evaluate the reliability of the annotation process, inter-annotator agreement scores were calculated. These scores provide insights into the consistency among annotators and the overall quality of the annotations, considering that sentences were presented without contextual information. We assessed inter-annotator agreement using three commonly employed metrics: Fleiss' Kappa, Cohen's Kappa, and Percentage Agreement.

\paragraph{Fleiss' Kappa.} Fleiss' Kappa is a measure of agreement between multiple annotators when categorizing items into multiple categories. For our task of classifying sentences as sexual (1) or non-sexual (0), Fleiss' Kappa was calculated as 0.79. This indicates substantial agreement among annotators.

\paragraph{Cohen's Kappa.} Cohen's Kappa measures the agreement between two annotators. It was used to assess pairwise agreement among our annotators. The average Cohen's Kappa across all pairs of annotators was found to be 0.72, indicating substantial agreement between individual pairs.

\paragraph{Percentage Agreement.} Percentage agreement, which measures the proportion of sentences for which all annotators agreed on the same label, was 87\%.

\subsection{Examples}
To illustrate the nature of sentences classified as sexual or non-sexual, we provide the following examples from our dataset along with their corresponding annotations in the Table \ref{tab:sentence-classification}.

\section{Sources of data}
All non-professional stories were scrapped from publicly available websites, including
\begin{itemize}
    \item \textit{opowiadaniaerotyczne-darmowo.com}
    \item \textit{sexopowiadania.pl}
    \item \textit{pornzone.com}
    \item \textit{anonserek.pl}
    \item \textit{opowi.pl} (categories: o życiu, różne, miłosne)
    \item \textit{opowiadania.pl}
    \item \textit{polki.pl}
\end{itemize}
We utilized maximum of 2 stories from the same author.

\newpage
\section{Details regarding training parameters}\label{app:parameters}

In this section we present parameters used for training baselines models (Table \ref{tab:params_baseline}) and classification reports on the baselines models (Tables \ref{tab:roberta10}, \ref{tab:roberta20}, \ref{tab:longformer}).

\begin{table}[ht]
  \caption{Classification Raport for validation set for baseline RoBERTa model trained on 10 epochs. Prec. means precision and Rec. means recall.}
  \label{tab:roberta10}
  \begin{tabularx}{\columnwidth}{Xcccc}
    \toprule
     & \textbf{Prec.} & \textbf{Rec.} & \textbf{F1} & \textbf{Sup.} \\
    \textbf{Non-harmful} & 0.96 & 0.71 & 0.82 & 38 \\
    \textbf{Harmful} & 0.78 & 0.97 & 0.87 & 40 \\
    \text{Accuracy} &  &  & 0.85 & 78 \\
    \text{Macro avg} & 0.87 & 0.84 & 0.84 & 78 \\
    \text{Weighted avg} & 0.87 & 0.85 & 0.84 & 78 \\
    \bottomrule
  \end{tabularx}
\end{table}

\begin{table}[ht]
  \caption{Classification Raport for validation set for baseline RoBERTa model trained on 20 epochs. Prec. means precision and Rec. means recall.}
  \label{tab:roberta20}
  \begin{tabularx}{\columnwidth}{Xcccc}
    \toprule
     & \textbf{Prec.} & \textbf{Rec.} & \textbf{F1} & \textbf{Sup.} \\
    \textbf{Non-harmful} & 0.75 & 1.0 & 0.85 & 38 \\
    \textbf{Harmful} & 1.0 & 0.68 & 0.81 & 40 \\
    \text{Accuracy} &  &  & 0.83 & 78 \\
    \text{Macro avg} & 0.87 & 0.84 & 0.83 & 78 \\
    \text{Weighted avg} & 0.88 & 0.83 & 0.83 & 78 \\
    \bottomrule
  \end{tabularx}
\end{table}

\begin{table}[ht]
  \caption{Classification Raport for validation set for baseline Longformer model. Prec. means precision and Rec. means recall.}
  \label{tab:longformer}
  \begin{tabularx}{\columnwidth}{Xcccc}
    \toprule
     & \textbf{Prec.} & \textbf{Rec.} & \textbf{F1} & \textbf{Sup.} \\
    \textbf{Non-harmful} & 0.82 & 0.89 & 0.85 & 35 \\
    \textbf{Harmful} & 0.90 & 0.84 & 0.87 & 43 \\
    \text{Accuracy} &  &  & 0.86 & 78 \\
    \text{Macro avg} & 0.86 & 0.86 & 0.86 & 78 \\
    \text{Weighted avg} & 0.86 & 0.86 & 0.86 & 78 \\
    \bottomrule
  \end{tabularx}
\end{table}

\begin{table*}[ht]
    \centering
    \caption{Examples of Sentence Classification with Sexual Content (A1 -- Annotator 1, A2 - Annotator 2, A3 -- Annotator 3) \textbf{Warning: This table contains sentences with sexual content. Reader discretion is strongly advised.}}
    \label{tab:sentence-classification}
    \begin{tabularx}{\linewidth}{|X|X|c|c|c|c|} 
        \hline
        Sentence & Translation & A1 & A2 & A3 & Majority Label \\
        \hline
        Pocałunek trwał kilka sekund. & The kiss lasted for a few seconds. & 0 & 0  & 0  & 0 \\
        \hline
        Obciągnęłam spódnicę i cofnęłam nogę pod stół. & I pulled down my skirt and withdrew my leg under the table. & 0  & 0  & 1  & 0\\
        \hline
        Marcin pomógł jej pozbyć się swoich spodni. & Marcin helped her get rid of her pants. & 0  & 1 & 0  & 0 \\
        \hline
        Robił to powoli, z czasem przyspieszył, a ja już nie mogłam. & He did it slowly, and over time he sped up, and I couldn't do it anymore. & 0 & 1 & 1 & 1 \\
        \hline
        Wstał, a ja uklęknęłam przed nim, wzięłam znowu do ust. & He stood up, and I knelt in front of him, took to my mouth again. & 1 & 1 & 1 & 1  \\ \hline
        Po chwili przyciągnął Joannę do siebie. & After a while, he pulled Joanna close to him. & 0 & 0 & 0 & 0\\ 
        \hline
    \end{tabularx}
\end{table*}

\begin{table*}
    \centering
    \caption{Training parameters for baseline models}
    \label{tab:params_baseline}
    \begin{tabular}{|c|c|c|c|}
    \hline
        Parameter & RoBERTa based 10 & RoBERTa base 20 & Longformer\\ \hline
        learning rate & $1e-5$ & $1e-5$ & $1e-5$ \\ \hline
        number of epochs & $10$ & $10$ & $10$\\ \hline
        optimizer & Adam & Adam & Adam \\ \hline
        batch size & $8$ & $8$ & $8$\\ \hline
        number of linear layers & $3$ & $3$ & $3$\\ \hline
        dropout probability & $0.2$ & $0.2$ & $0.2$\\ \hline
        activation layer & ReLU & ReLU & ReLU\\ \hline
    \end{tabular}
\end{table*}

\end{document}